\documentclass[a4paper, 11 pt]{article}
 \pdfoutput=1

\usepackage{geometry}
\usepackage[title]{appendix}
\usepackage{amssymb}
\usepackage{amsmath}
\usepackage{placeins} 
\usepackage{graphicx}
\usepackage{epsfig}
\usepackage{multirow}
\usepackage{subfigure}
\usepackage{authblk}

\usepackage[ruled]{algorithm2e}
\usepackage[dvipsnames]{xcolor}
\usepackage{enumitem}
\newlist{abbrv}{itemize}{1}
\setlist[abbrv,1]{label=,labelwidth=1in,align=parleft,itemsep=0.1\baselineskip,leftmargin=!}
\usepackage{hyperref}
\usepackage{textcomp}
\usepackage{hyperref}

\title{\textbf{ On-board Fault Diagnosis of a Laboratory Mini SR-30 Gas Turbine Engine}}

\author[1]{Richa Singh\footnote{Corresponding author - Richa Singh/richachauhan92@gmail.com}}
\affil[1]{\textit{Department of Aerospace Engineering Indian Institute of Technology Bombay, Mumbai - 400076, India}}

\begin{document}
	
	\date{}
	
	\maketitle 
	\thispagestyle{empty}
	
\vspace{-1 cm}	
	\begin{center}
		\section*{ Abstract}
	\end{center} 
		Inspired by recent progress in machine learning, a data-driven fault diagnosis and isolation (FDI) scheme is explicitly developed for failure in the fuel supply system and sensor measurements of the laboratory gas turbine system. A passive approach of fault diagnosis is implemented where a model is trained using machine learning classifiers to detect a given set of fault scenarios in real-time on which it is trained.  Towards the end, a comparative study is presented for well-known classification techniques, namely Support vector classifier, linear discriminant analysis, K-neighbor, and decision trees. Several simulation studies were carried out to demonstrate and illustrate the proposed fault diagnosis scheme's advantages, capabilities, and performance.	\\

\textbf{Keywords:}	Confusion matrix, Gas Turbine Engine,  Fault diagnosis, .

	\section{Introduction}
	
	Fault diagnosis plays a crucial role in condition monitoring and health management. It can avoid major accidents, significant economic losses, and the cost associated with untimely and unnecessary replacement of the components. It also extends the service life of the machines and decreases maintenance costs by avoiding unnecessary shutdowns. Gas turbine fault detection is essential for minimizing operational costs related to the overhaul time intervals. Thus, it is vital to perform fault diagnoses regularly to maintain the reliability of the gas turbine engine. Early diagnosis of anomalies and faults in an engine makes it possible to perform crucial condition-based maintenance decisions and actions. However, fault diagnosis is still a challenging area because of the increased complexity of
the system day by day, and hence coming up with an efficient fault diagnosis technique that is applicable in the online operation of power plants involved with minor computations is an urgent need in power generation industries \cite{khan2018review}.

The safety, reliability, and performance of complex systems incorporated with many sensors depend on the sensors' accuracy. Sensor readings mainly determine the system state. The denotation of an abnormal state may be the consequence of a sensor failure rather than a system flaw. 	In some cases, failure to identify the source of an ``abnormal state'' and take corrective action may result in expensive and unnecessary system shutdowns or accidents. Sensor readings are also used for feedback control purposes, where this could lead to inaccurate control. Hence, it is essential for a monitoring and diagnostic system between the case where a sensor failure and not a system fault is responsible for the indication of an abnormal state. The guarantee ensured by FDI leads to a systematic way to meet the performance specifications of control systems in industrial applications that can be performed in an optimized way. 

Fault diagnosis techniques can be commonly categorized as model-based and data-driven approaches. Both techniques
have been extensively studied in the literature for the health monitoring of complex nonlinear systems. In model-based techniques, Kalman filters are quite popular \cite{kobayashi2003application}. In online parameter estimation, which contributes to the system degradation. The model-based technique has been explored to identified slow varying/evolving fault propagation in the gas turbine by implementing the Kalman filter, and its derivative method \cite{kobayashi2003application, orchard2007particle, hanlon2000multiple, willsky1974generalized}. Model-based techniques have advantages in terms of on-board implementation considerations but are restricted to linearized models. Moreover, their reliability often decreases as the system's nonlinear complexities and model uncertainties increase due to their dependency on the mathematical model and sensor information. In addition to that, the main drawback is that it fails to predict abrupt degradation in the system.

During the past few years, data-driven techniques which rely on real-time data from the experimental setup has gained interest due to the availability of massive data set made possible by improved sensor quality and increase in computation power. In that series, deep neural networks do not require a detailed numerical model of the aircraft engine and provide flexible and reliable tools for dealing with nonlinear problems and modeling complex and nonlinear dynamical systems with excellent capability. In \cite{vanini2014fault, kiakojoori2016dynamic} FDI scheme is presented to detect and isolate faults in a highly nonlinear dynamics of an aircraft jet engine consist of multiple DNNs or parallel bank of filters, corresponding to various operating modes of the healthy and faulty engine conditions. However, training multiple DNN requires much computation. The accuracy of such a model relies on trial and error analysis. Similar results can be achieved with a shallow network that requires less computation, such as presented in \cite{widodo2007support, muralidharan2012comparative}. It is also seen that Support vector machine, logistic regression-based classifiers are not suitable for multi-class classification. This method comes under shallow machine learning algorithms. In this work, a Linear discriminant analysis-based classifier \cite{balakrishnama1998linear} is chosen as the FDI Scheme as it comes with an advantage of multi-class classification extension.

A data-driven fault diagnosis scheme is developed specifically for the fuel supply system and sensors of a laboratory GTE. The proposed FDI Scheme is constructed using only the system I/O data at different engine operating points. The FDI scheme works on a passive approach where the model is trained for a specific set of faults and utilized further in online detection. Multiple model approach is used for fault isolation which means a separate classifier will correspond to a specific fault. The work is later extended by performing fault diagnosis in a gas turbine engine which is considered a classification problem where the component status is predicted as either healthy or faulty. The proposed fault diagnosis model is designed to capture the two main degradations in the gas turbine engine, namely the fault in the fuel supply system and sensor failure using Machine Learning (ML) classifiers.  The classification performance of real-time data is compared with well-known machine learning classifier namely Support vector machine (SVM) \cite{awad2015support}, k-neighbour (KNN) \cite{zhang2005k}, linear discriminant analysis (LDA)~\cite{balakrishnama1998linear}, and decision tree~\cite{priyam2013comparative}. The proposed approach does not require any prior knowledge of the mathematical model, which is an important advantage over the model-based techniques. Moreover, for different fault scenarios in the system, the best of the available classifiers is deployed in real-time to deal with on-board fault detection to enhance the accuracy rather than relying on a single classifier. The proposed technique can be used for health monitoring of the engine, specifically for sensor validation. 

It is essential to point out that faulty data collection is not feasible with the actual aircraft engine. Thus,  several works of literature exploit high fidelity mathematical model which integrates the thermodynamic propulsion system in MATLAB/Simulink with varying levels of difficulty. These tools include Numerical Propulsion System Simulation (NPSS), T-MAPS, GasTurb TurboLib, and Commercial Modular Aero-Propulsion System Simulation (C-MAPSS), available as open-source to industry professionals or academia. In the current study, a First-principle based mathematical model of the engine is developed by using the state variable method. The detailed description of modeling of laboratory GTE and fuel supply system can be referred to in \cite{Turbine}. The model is in good agreement and gives satisfactory performance when validated against the experimental runs.

The paper is organized as follows: Section \ref{sys_des} gives a brief description of laboratory gas turbine engine. It also discusses the problem formulation for fault diagnosis for the current work. Section \ref{FDI_2} gives an insight into fault diagnosis procedure, method, and state of the art of introducing a fault in the system, followed by the results and discussions in Section~\ref{result}. Lastly, Section \ref{Conc} concludes the paper and outlines future research.

	\section{System Description}
	\label{sys_des}
The Turbine Technologies, LTD MiniLab Gas Turbine Power System is a complete, self-contained jet engine laboratory featuring the purpose-built SR-30 Turbojet Engine. Designed expressly for engineering education and research purposes, the MiniLab allows all aspects of gas turbine theory to be easily demonstrated and readily explored. A pure turbojet, the SR-30 engine is representative of all straight jet engines in which combustion results in an expanding gas that is sufficiently capable of producing useful work and propulsive thrust. The SR-30 engine is typical of the gas generator found in turbofan, turboprop, and turboshaft gas turbine engines.  The miniLab engine is equipped with (a) centrifugal flow compressor, (b) reverse-flow combustion chamber, (c)  axial flow power turbine, and (d) exhaust nozzle.  These types of engines are used for aircraft, defense systems, and marine propulsion, as well as stationary and industrial power generation~\cite{Turbine}.

\begin{figure}[ht!]
	\centering
	\subfigure[]{\includegraphics[width = 9 cm]{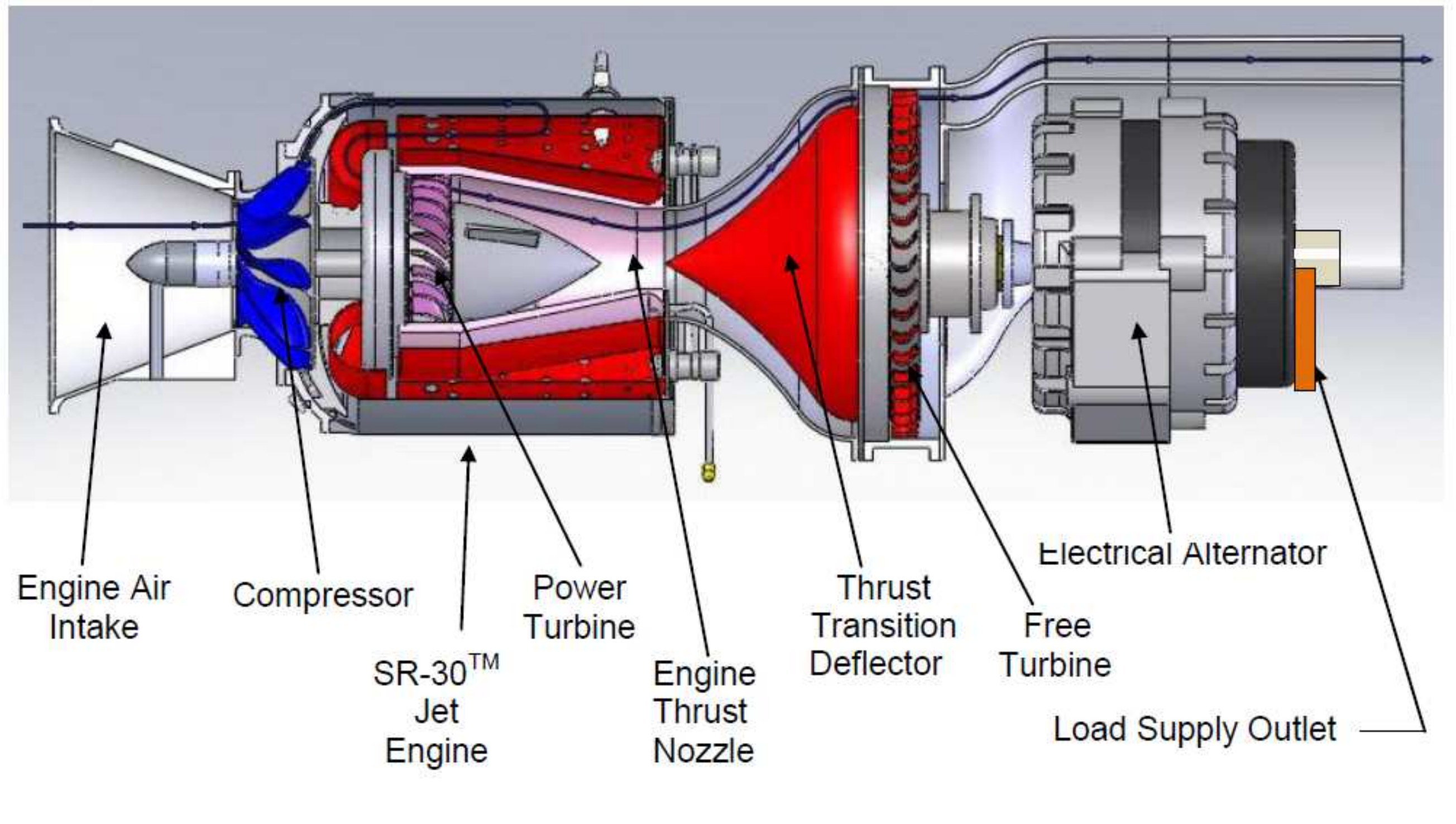}\label{sr30}} 
	\subfigure[]{\includegraphics[width = 9 cm]{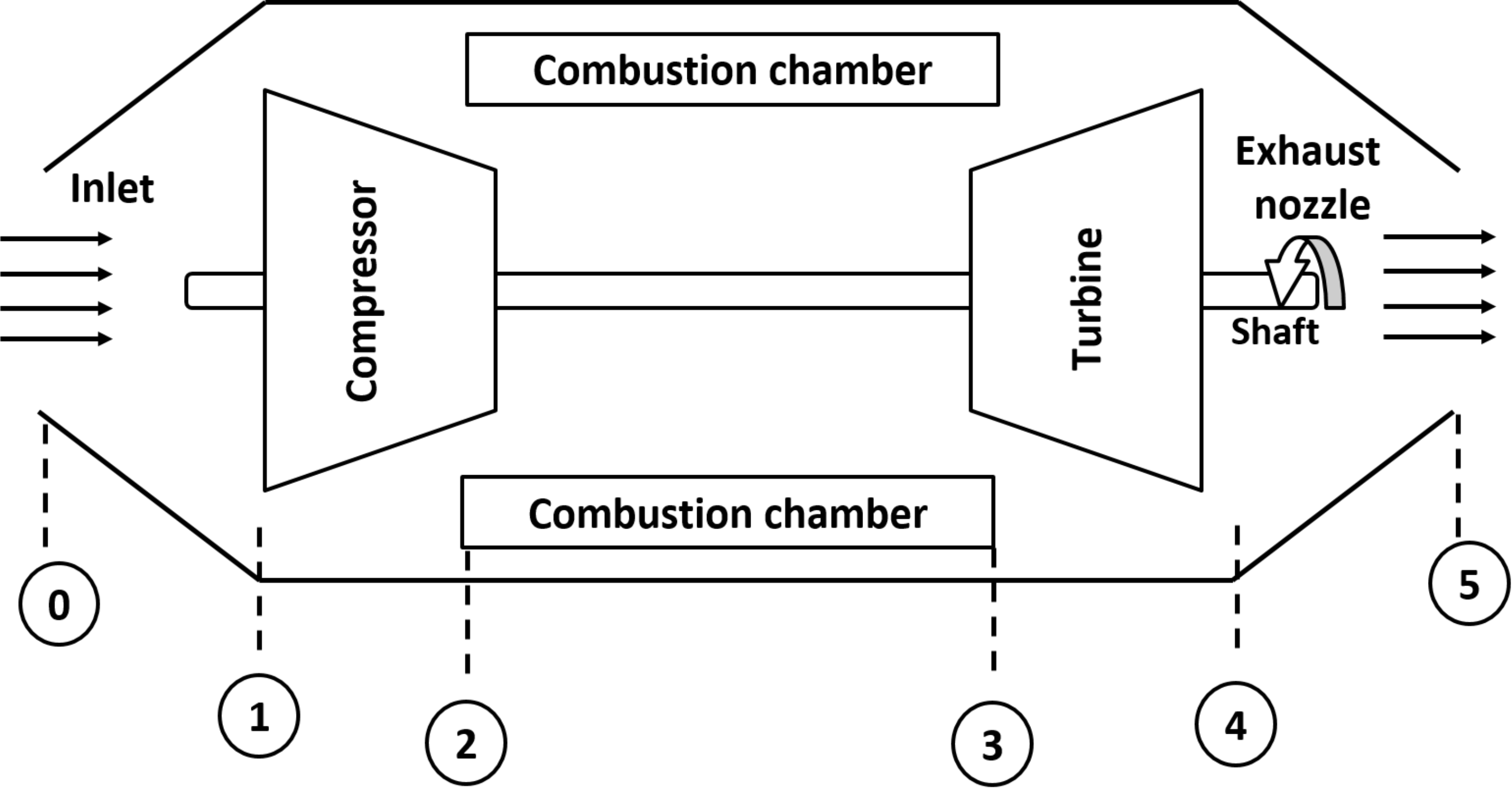} \label{schematic_sr30}} 
	\caption{Laboratory mini SR-30 GTE \subref{sr30} cross-sectional view \subref{schematic_sr30} schematic representation of engine components}
	\label{Fig:sr30}	
\end{figure}

Theoretical predictions can be measured on actual hardware utilizing the included data acquisition system. Sensors located along the gas flow path allow accurate, real-time measurements of the operating conditions at these points. The MiniLab DigiDAQ System utilizes an IOtech Personal Daq/56 USB Data Acquisition Module.
Featuring 22-bit analog to digital conversion, multiple channels of voltage, thermocouples, pulse, frequency, and digital I/O can be measured and controlled. This is accomplished through 20 single-ended or ten differential analog (up to $ \pm$20V full scale) or thermocouple input channels, 16 programmable ranges, 500V optical isolation, 16 digital I/O lines, and four frequency/pulse channels. The integrated USB connection allows a single cable interface of up to 16 feet (5 meters) between the MiniLab and the data acquisition computer. This distance quickly increases to 98 feet (30 meters) using powered USB hubs (serving as data repeaters). The USB’s high-speed data transfer rate (up to 12Mbits/s) allows for a real-time display of acquired data while eliminating the need for buffer memory in the data acquisition system itself.

Fig 2 shows the station index numbers, which have been followed in this paper. In the current scenario, the fuel flow $(\dot{m}_{f}) $ is the control input to the engine, which is regulated to the combustion chamber through a servomechanism device called servo actuator. The output parameter are the pressure $ (P_{i}) $, temperature $ (T_{i}) $ at each section $ (i) $  as represented in Fig~\ref{Fig:sr30}(b) and shaft speed $ (N) $  of rotor. Since the engine is contained in minilab, the ambient parameter such as ambient pressure $  (P_{1})$, ambient temperature $ (T_{1}) $, mass flow of air through inlet $ (\dot{m}_{a}) $ considered to be constant. The reference value of parameter at ideal operating range is given below  in Table \ref{Table:reference_value}. 

\begin{table}[ht!]
	\begin{center}
		\small
		\caption{The reference value specifications of the engine model \cite{Turbine}}
		\label{Table:reference_value} 
		\resizebox{\textwidth}{!}{ 
			\begin{tabular}{c l c c }
				\hline 	{\textbf{Symbol}} & \textbf{Meaning}  & \textbf{Unit}  &\textbf{ Design point value} \\
				\hline
				\hline
				\multicolumn{4}{l}{\textbf{Reference  Value and constants}}\\ 
				\hline
				$ P_1, T_1 $ &	Ambient pressure and temperature at inlet&	kPa, K &	101 kPa, 300 K\\			
				$ \dot{m}_{ref} $	& Mass flow rate of air	&kg/sec&	0.5\\
				$ \gamma $&	Heat capacity ratio	& - &	1.4\\
				$ 	LHV $&	Lower heating value of fuel&	kJ/kg-K	& 43000 \\
				$ c_p $&	Specific heat of mixed gas at constant pressure &	kJ/kg-K &	1.075\\
				$ 	R $	&Characteristic gas constant& 	kJ/kg-K	& 0.287 \\
				$ I $&	Mass moment of inertia of the shaft& 	Kg-m2&	$3.2 \times10^{-4}$ \\				
				$\sigma_{cc}$ & 	Combustion chamber pressure loss & - & 0.96\\
				$ 	V_1, V_2 $&	Pseudo-volume generate  between components during transient operation&	$ m^{3} $& 	0.24, 0.36 \\
				\hline  	
				\multicolumn{4}{l}{\textbf{Parameter to be calculated through simulation}}\\
				\hline
				$ 	P_2, T_2 $	& Compressor exit pressure and temperature &	kPa, K&	\\
				$ P_3, T_3 $&	Turbine inlet pressure and temperature &	kPa, K&	  \\ 
				$ 	P_4, T_4 $&	Turbine exit pressure and temperature 	&kPa, K	&  \\
				$ P_5, T_5 $&	Exhaust nozzle pressure and temperature& 	kPa, K&	 \\
				$ 	N $	&Shaft speed 	& rpm &	\\
				$ \dot{m}_{f} $	& Fuel flow rate	&L/Hr&	\\
				$ \eta_c , \eta_t $	&Compressor and turbine efficiency& 	-&\\
				$ 	W_{c}, W_{t} $ &	Work consumed by compressor and produced by turbine	&kJ/kg-s&	\\
				$\dot{m}_c,\dot{m}_t,\dot{m}_n $&
				Mass flow rate through compressor, turbine, and nozzle &	Kg/sec	&  \\			
				\hline
		\end{tabular} }
	\end{center}
\end{table}

\subsection{Modeling of the Laboratory SR-30 Engine} 
\begin{figure}[ht!]
	\centering
	\includegraphics[width = 14 cm]{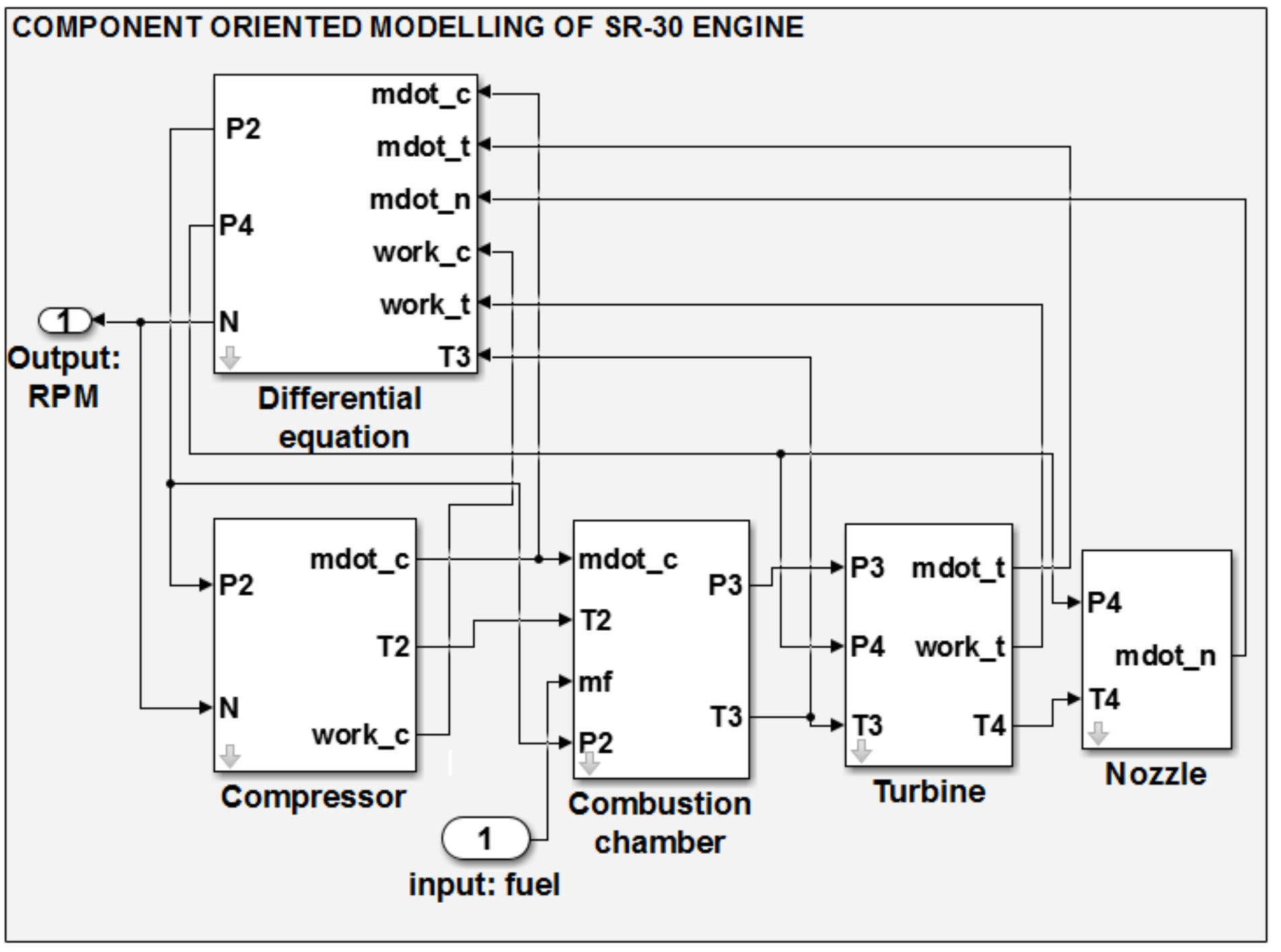} 
	\caption{Station index numbers of the SR-30 gas turbine engine}
	\label{Fig:simulink_model}
\end{figure}

\begin{figure}[ht!]
	\centering
	\subfigure[]{\includegraphics[scale=0.5]{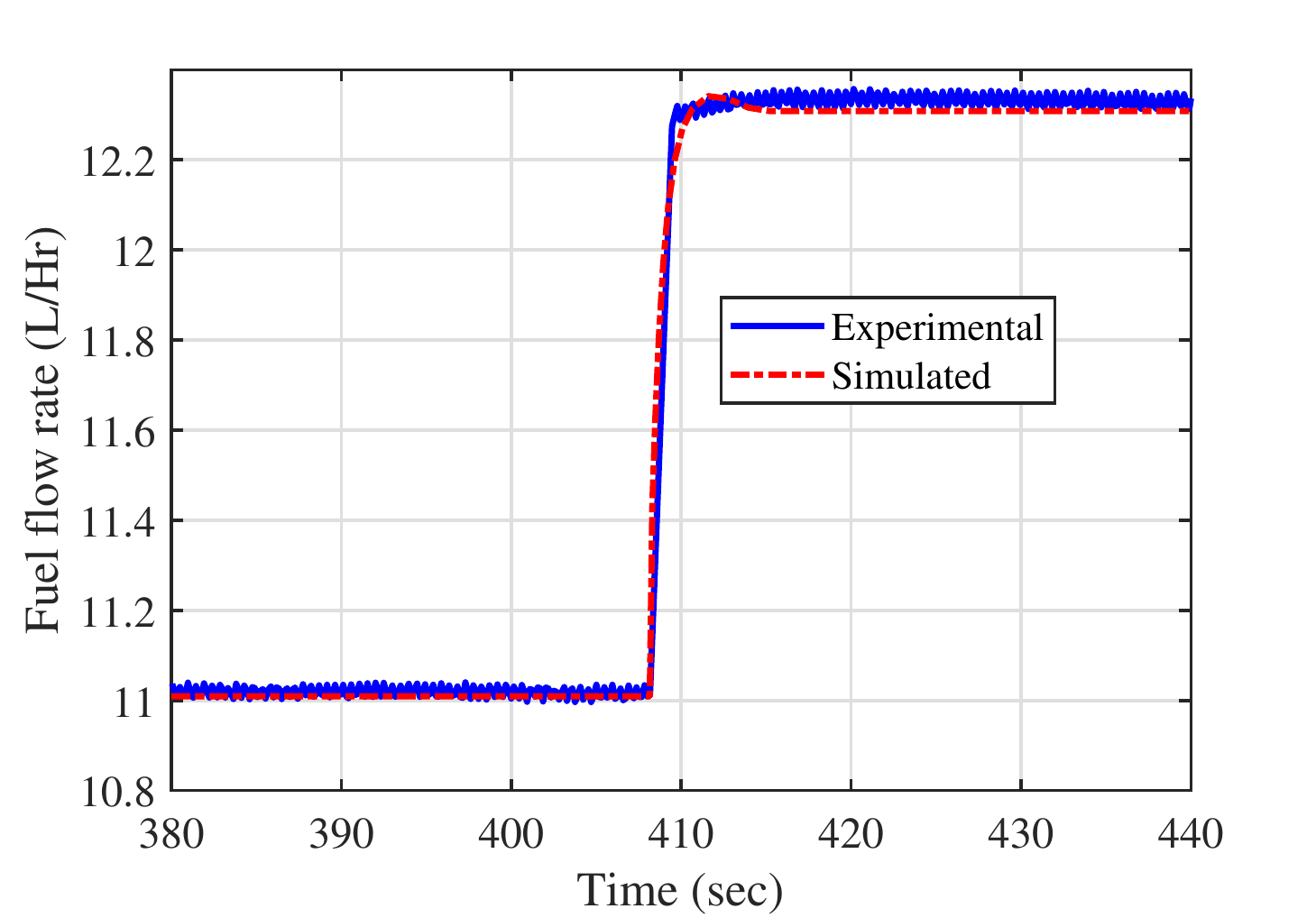}    \label{fuel_val}} 	
	\subfigure[]{ \includegraphics[scale=0.45]{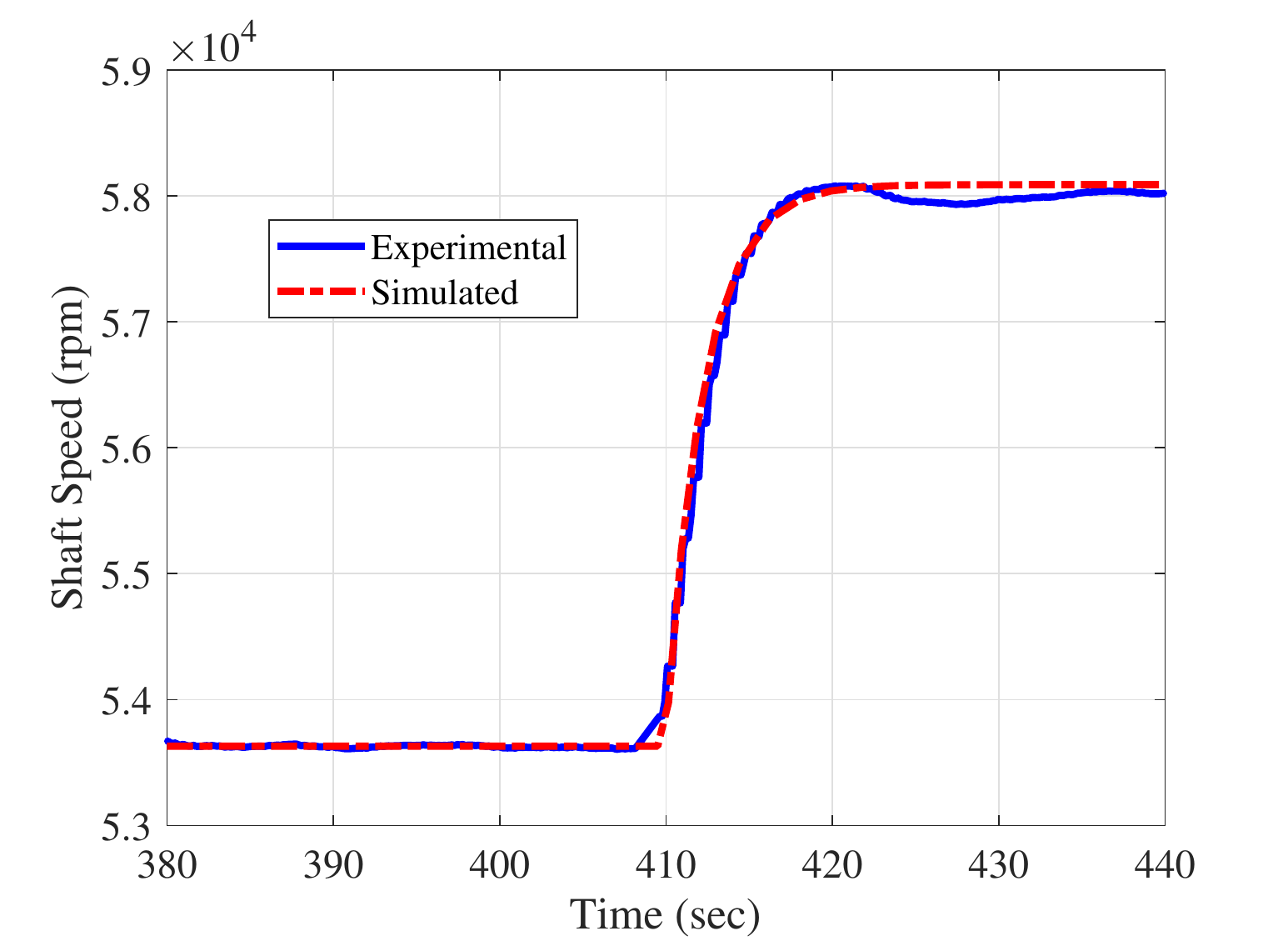}    \label{rpm_val}} 
	\caption{GTE model augmented with actuator dynamics validation with experimental runs  \subref{fuel_val}  control effort $ (\dot{m}_{f}) $, and \subref{rpm_val}  shaft speed $ (N) $}
	\label{model_val}
\end{figure}

The modeling approach presented here begins with a proper selection of state variables that are sufficient to derive all the governing equations and engine parameters independently. The state variables are selected such that, for the given external conditions and inputs, the performance of each component can be evaluated with one pass of an engine calculation \cite{phdthesis}.
Based on the engine physics, a component-wise mathematical model of laboratory mini SR-30 GTE  is developed using the \textit{State variable method} is modeled using a set of three ordinary differential equations, which can be defined by associating with each state variable, i.e., $ [P_{2}, ~P_{4}, N] $. The  can be described by

\begin{align} \label{Eq.:P2_dot_SS_1}
	\dfrac{dP_{2}}{dt} &= \dfrac{R}{V_1}\left[ (\dot{m}_c - \dot{m}_t) T_2 + \dot{m}_c\dfrac{dT_{2}}{dt} \right] \\
	\label{Eq.:P4_dot_SS_1}
	\dfrac{dP_{4}}{dt} &= \dfrac{R}{V_2}\left[ (\dot{m}_t - \dot{m}_n) T_4 + \dot{m}_t\dfrac{dT_{4}}{dt} \right] \\
	\label{Eq.:N_dot_SS_1}
	\dfrac{dN}{dt} &= \dfrac{1}{N}\left( \dfrac{60}{2 \pi} \right)^2 \frac{W_t - W_c}{I} 
\end{align}

where $ R $ is characteristic gas constant, $ V_{1}, V_{2}$ are the pseudo-volume created between engine components in transient process, $ I $ is moment of inertia of shaft, and $ \dot{m}_{c} $,  $ \dot{m}_{t} $ and  $ \dot{m}_{n} $ are the mass flow rate of compressor, turbine and nozzle respectively. The work input to the compressor $  W_{c}  $ and work produce by turbine is $ W_{t} $.   $ P_{i} $ and $ T_{i}$ are the pressure and temperature respectively at $ i $th station index with respect to Fig. \ref{Fig:simulink_model}. 
The mathematical simulation model of GTE is validated against experimental runs. The fuel flow rate from the experimental dataset is fed to the simulation model in the form of a step waveform, and the corresponding shaft speed responses predicted by the GTE simulation model are logged as shown in Fig. \ref{model_val}. 

The model is found to deliver good shaft speed prediction with an accuracy of 88.2\%.	The detailed description of this model and its validation is given in \cite{Richa}.  This virtual model is utilized to simulate different fault scenarios in the current work as it is not feasible to create faults in the laboratory engine. 
 
\subsection{Modeling of the fuel supply system} 
The actuator dynamics is also incorporated here along with the physics-based model, which is termed as fuel supply system (FSS). The FSS of a gas turbine engine includes a servo actuator, throttling lever, fuel pump, and fuel injector. A servo-actuator, data acquisition system, and software have been implemented in the laboratory SR-30 engine to eliminate the manual operation of fuel flow via throttling lever. This throttling lever is mechanically linked with a fuel pump which opens to a reservoir. The reservoir fuel injector allows the fuel flow to the combustion chamber of the engine. An independent actuator dynamic for the servo control mechanism, which regulates the fuel flow to the combustion chamber with the pulse signal, is incorporated. The angle measures the fuel flow moved by the throttle lever with respect to the input signal. 
\begin{figure}[ht]
	\centering
	\includegraphics[scale = 0.6]{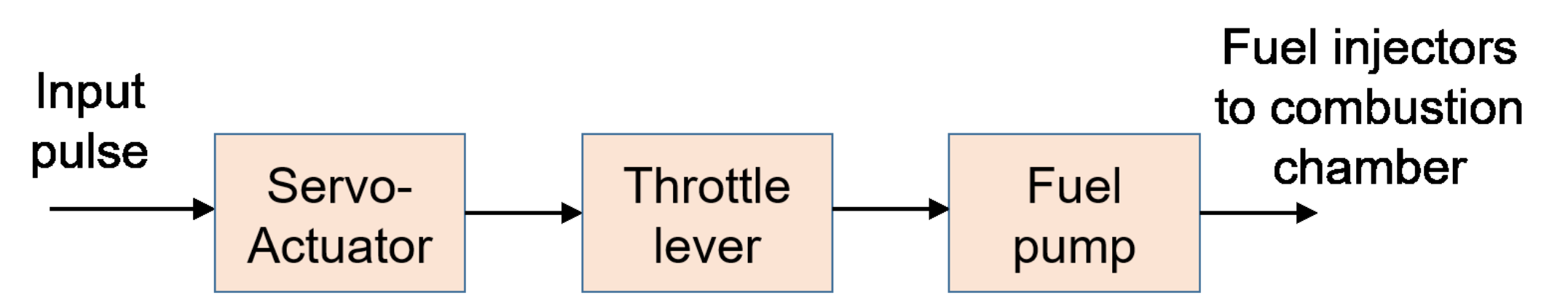} 
	\caption{Fuel supply system of gas turbine engine}
	\label{Fig:fss}
	
\end{figure}

Nonlinear autoregressive networks with exogenous inputs (NARX) have been adopted in the modeling of fuel supply systems as they are highly potential in capturing the dynamics of nonlinear systems with single-input and single-output (SISO).  A large amount of data is collected at various operating ranges to be applied to the networks to learn its dynamics. A network with two hidden layers is required to learn the dynamics of the fuel system. NARX neural network with two hidden layers having 10 and 7 neurons respectively with   \textit{Levenberg-Marquardt} function as optimizer \cite{ranganathan2004levenberg}, is used to model fuel supply system. 

\begin{figure}[ht]
	\centering
	\includegraphics[scale=0.5]{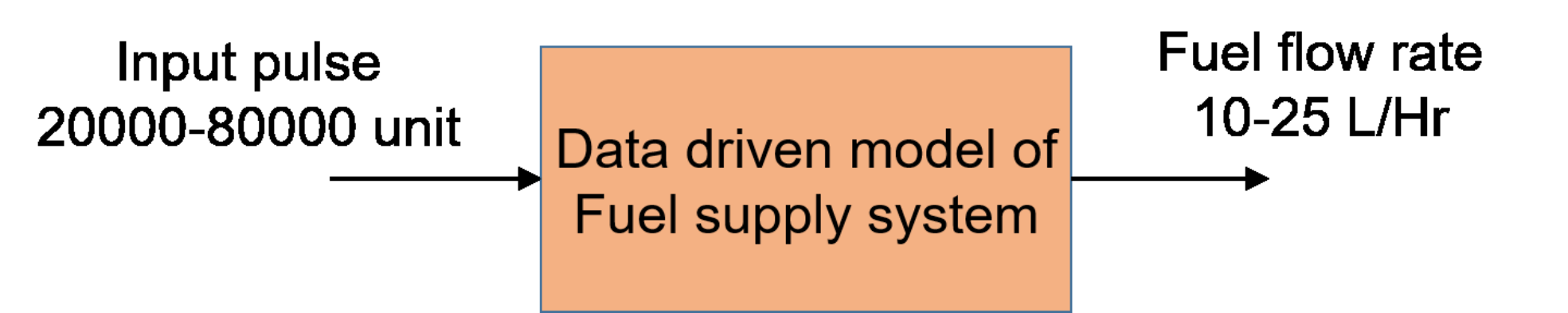} 
	\caption{Deep Neural network for Fuel supply system}
	\label{Fig:fdd}
\end{figure}

The developed NARX model of FSS is validated against experimental runs, where a step signal in the pulse is given as input to the network, and correspondingly the predicted fuel flow rate response is logged.

\begin{figure*}[h!]
	\centering
	\subfigure[]{\includegraphics[width=6.5 cm]{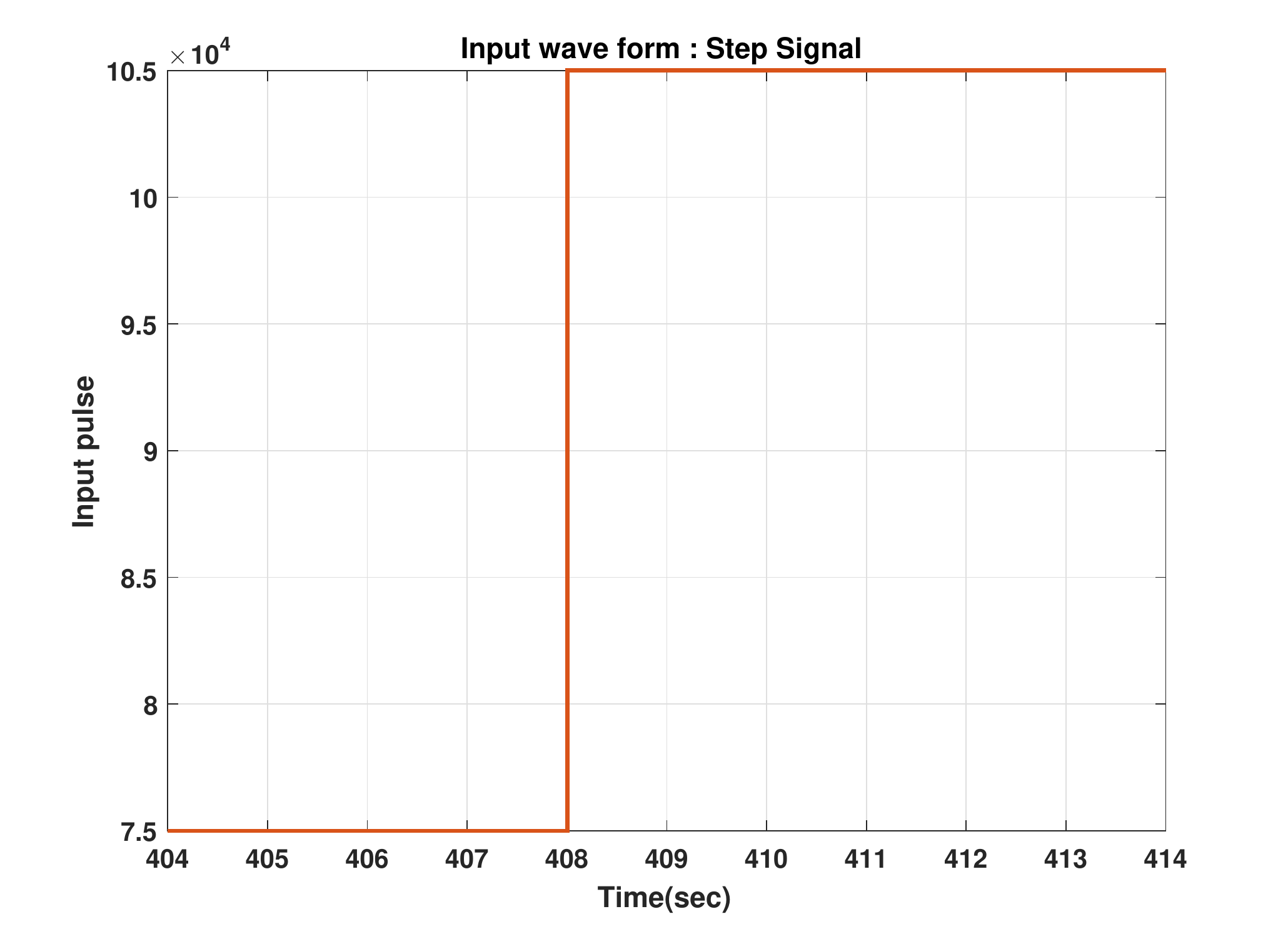}\label{pulse_imput}}
	\vspace{-0.4 cm}
	\subfigure[]{\includegraphics[width=6.5 cm]{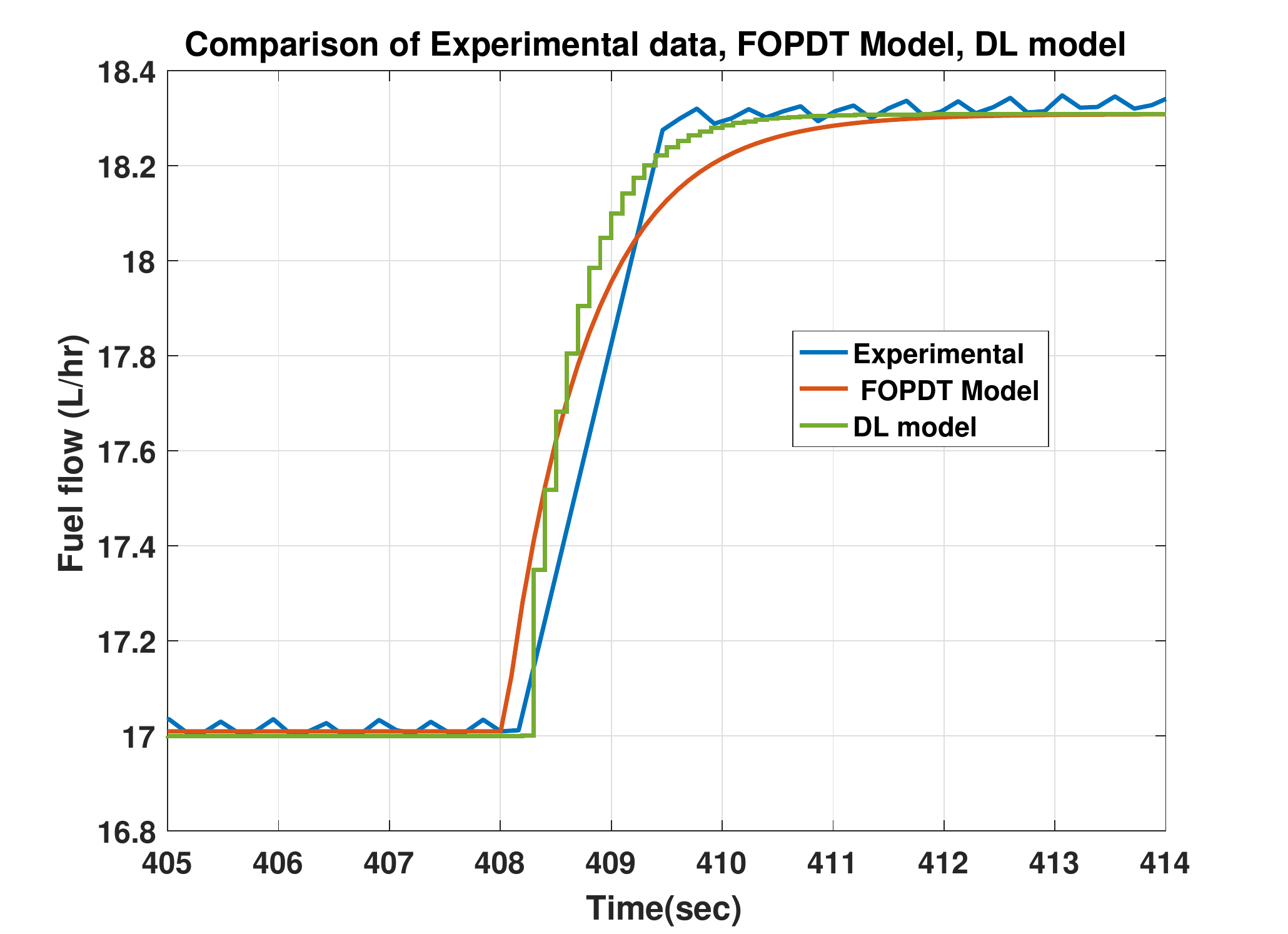}\label{fuel_output}}
	\caption{Fuel supply system validation: \subref{pulse_imput} Pulse input and \subref{fuel_output} Predicted fuel flow rate}
	\label{Fig:FSS_Validation}	
\end{figure*} 
Figure~\ref{Fig:FSS_Validation}(b) shows that the prediction result obtained from the NARX model follows the experimental data. The model output is also verified against a first order with dead time (FOPDT) model as given in \cite{Richa}. The NARX model of the fuel supply system predicts the fuel flow rate with an 88.9 \% accuracy. The detailed description of FSS modeling is given in \cite{Richa}.

\section{Proposed Fault Diagnosis procedure}
	\label{FDI_2}
	
The fault diagnosis problem in a gas turbine engine can be considered as a classification problem where the goal is to identify the current status of the engine components as healthy or faulty. The classification learner application in MATLAB allows supervised learning tasks such as interactively exploring the data, selecting features, and trained models using various classifiers by learning through the provided training set. The raw code can be extracted in MATLAB to use	the trained classifier in on-board diagnosis \cite{Matlab}. In this study, four different machine learning classifiers developed specifically Linear discriminant analysis (LDA), Support vector machine (SVM), K-neighbor, Decision tree, and their performance is compared for on-board fault detection in the laboratory gas turbine engine.  

The FD procedure can be precisely completed in three steps which are as follows:

\subsection{Benchmarking Datasets and preprocessing} 
\label{sec:data_collection}
An in-house,  physics-based model\cite{Richa} is  developed in Simulink/ MATLAB \cite{Matlab} is used to simulate  faults in a gas turbine engine. This mathematical model of laboratory GTE also allows identifying the characteristic parameters of the different components of the gas turbine, as listed in Table~\ref{Table:reference_value}. In present case study, out of available sensor measurement the following following parameters ($\dot{m}_{f}, ~P_{1}, ~T_{1}, ~P_{2}, ~T_{2}, ~P_{3}, ~T_{3}, P_{4}, ~T_{4}, P_{5},  ~T_{5},~N $) are selected in this study by the guide of naked eye examination of the degradation trends.

\begin{figure}[ht!]
	\centering\includegraphics[scale=0.6]{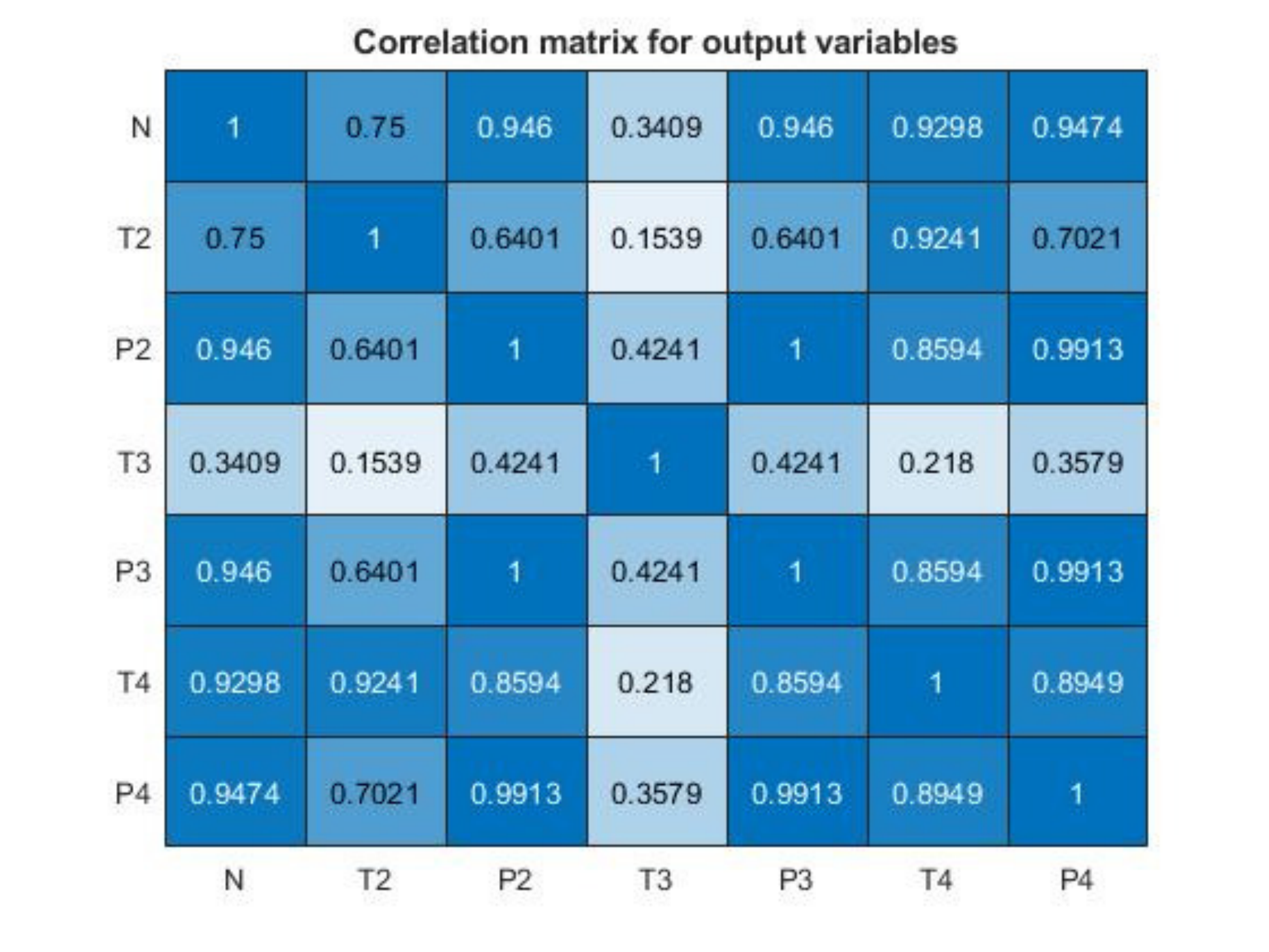}
	\caption{The correlation matrix representing the inter-dependency of measurements}
	\label{Fig:correlation_matrix}
\end{figure}

Various types of faults are created in the fuel supply system and output parameters such as temperature and pressure sensors. Sensor faults are created by amplifying or degrading the healthy sensor output at a different time instant.  The simulations are performed over a period of 100 seconds and involve 20 data sets with 1000 samples each for every type of fault studied. The data includes all monitored parameters and covers the entire operating range of the gas turbine engine. The data set has healthy as well as faulty samples, and hence the data is labeled accordingly. The healthy data is labeled as 0, and the erroneous data is labeled as 1. Further, $ \pm 2\% $ white  Gaussian noise is added to sensor measurement noise. The dataset is collected by manually introducing faults from the simulation model, as given in Table~\ref{Table:faults_data}.
In addition, the correlation matrix of the selected measurement is shown in Fig.~\ref{Fig:correlation_matrix}, which defines the inter-dependency of GTE measurements.  

\begin{table}[ht!] 
	\begin{center}
		\begin{small}
			{}			\caption{Description of the dataset considered for IFD of laboratory GTE}\label{Table:faults_data}
			\begin{tabular}{|l|l|l|}   
				\hline
				\textbf{Dataset} &\textbf{FD001}  &\textbf{FD002}   \\
				\hline	
				\textbf{Description} &  Actuator fault & Sensor failure     \\	 
				\textbf{Training set} & 10 &20  \\
				\textbf{Testing set} & 3 & 5  \\
				{\textbf{Fault mode}} & $ \begin{cases}
					F1 &   \text{Healthy}  \\
					F2 & \text{loack-in-place}\\
					F3 &  \text{bias}
				\end{cases} $	& $\begin{cases}
					F1 &   \text{Healthy}  \\
					F2 & \text{fault in} ~T_{2}\\
					F3 &  \text{fault in} ~T_{3}\\ 
					F4 &  \text{fault in} ~T_{5}\\
					F5 & \text{fault in} ~P_{2}
				\end{cases} $  \\
				\hline
			\end{tabular}
			\label{dataset_faulty}
		\end{small}
	\end{center}
\end{table}

\subsection{Data Preprocessing}
\label{sec:data_processing}
The collected data obtained from in-house simulator of GTE is preprocessed befor it used as training set in developing ML classifiers. As describe in section~\ref{sec:data_collection}, first a huge set of data is collected  for various fault scenarios. The data preprocessing includes (i) First we clean the data, filling the missing values and removing outliers from the data. Further we visualize the dataset to acess the trend, outliers and pattern of feature, to understand and eliminate the redundent features. (iii) Then, we identify the set features actually affecting the system performance. (iv) Next we label the data obtaied from simulation runs for different fault scenarios. Each data includes features and labels, where the features includes the time history of the parameters and label tells the status of the engine based on the features. (v) Since features includes parameters with different magnitudes, it is essential to bring all the featrue to a same scale magnitude so that the classifier is not biased. For this purpose we normalized the dataset, being used for training and testing the fault diagnosis classifiers.

\subsection{Training and Performance evaluation}

Before delving into the training process of classifiers, we need to divide the dataset into training and testing set. However, to evaluate the machine learning model offline, training training is to be performed on a limited dataset. Since the objective is classification, K-fold cross validation (CV) is preferred. This is generally used to estimate how the model is expected to perform in real-time. In k-fold cross validation, the original data is randomly divided into k subsets of equal sample size. During the training process, one of the subset is retained for validation and rest k-1 subset use for training in 1 epoch. in next epoch the other subset is kept for validation and rest are used for training classifier. 
The k-fold CV ensures that all the datasets are used for training and validation atleast once. The classification learner application  in MATLAB allows performing 5 fold cross-validation where  the training data is equally distributed into 5 subsets. The loss function is chosen as cross-entropy, as the target is categorical.

\begin{figure}[ht!]
	\centering
	\includegraphics[scale=0.6]{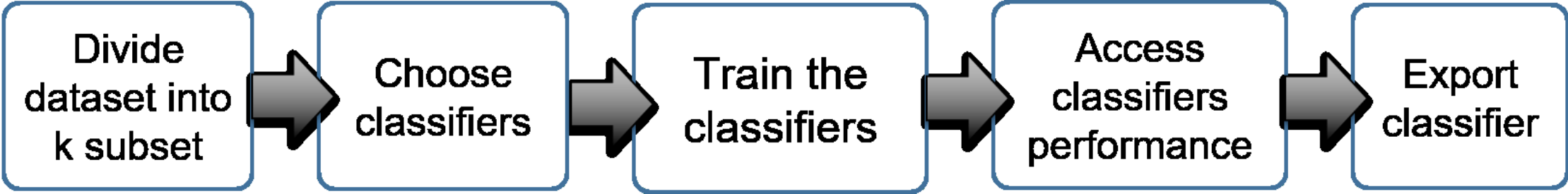}
	\caption{Flow chart shows a common workflow for training classifiers in the Classification Learner app.}
	\label{Fig:classifier_learner_app}
\end{figure}

Next using MATLAB's Classifier learner application toolbox, the machine learning classifier used for comparison purposes is trained.  Choosing the best classifier is left on the performance after training with all the available in-built classifiers. Specifically, four classifiers have been chosen primarily, and the training performance is evaluated for the various performance matrices such as accuracy, $ F1 $ score, and training time. While training for the different faulty datasets, the classifier's performance varies depending on the nature of the fault.  Figure~\ref{Fig:classifier_learner_app} shows the common workflow to develop a classifier from Matlab \textit{classifier learner application}. Once, the accurate classifier is selected, this can be extracted and can be used on edge devices to predict the status of a dynamical system.

\subsection{Deploy algorithm}

\begin{figure}[ht!]
	\centering
	\includegraphics[scale =  0.5]{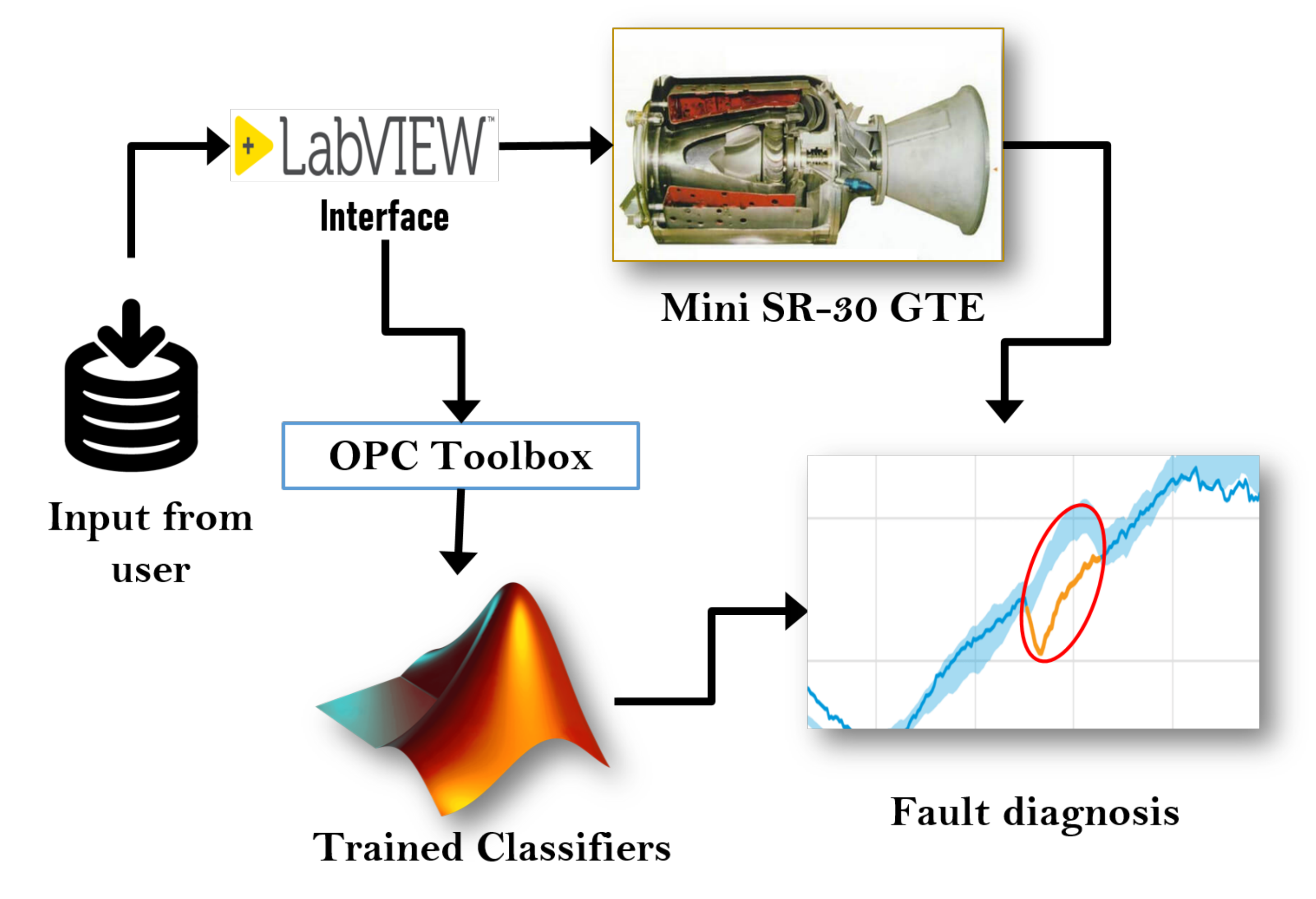} 
	\caption{Schematic flow of fault diagnosis procedure}
	\label{Fig:FDD deployment}
\end{figure}

Based on the performance evaluation, the classifier delivers the best performance among all is extracted and deployed in a real-time system for on-board fault detection to make predictions on new test data. Given the input, the faults created in a particular sensor or fuel supply system, and the data passed through the trained classifier model. The model classifies the sample as healthy/faulty depending upon the predictors, i.e., input to the classifier model. 
 It is to be pointed that the realtime data is logged in the LabView interface is read by MATLAB software via open platform communication (OPC). OPC toolbox provides access to live data directly to MATLAB from LabView.   This assembly is constituting the National Instrument Data acquisition system, LabView interface, and MATLAB Toolbox to perform hardware-in-loop model validation developed by the proposed deep learning framework.   Here, MATLAB serves as a client and collects live data, and fed as input to the trained classifier in MATLAB and engine health is monitored by on-board fault diagnosis as shown in Fig.~\ref{Fig:FDD deployment}.  

\section{Results and validation}
\label{result}
Multiple classifier models have been trained and deployed on-board, each corresponding to a specific engine and fuel supply system failure. The first principle-based simulator is used to generated data sets for healthy and various fault scenarios. The offline training is accomplished using a machine learning classifier in MATLAB/Simulink environment. The classifier which delivers better performance is then deployed on-board. The on-board performance is evaluated offline using the confusion matrix on the test dataset.

\textit{Note:} 
Table II- V shows the offline training performance comparison for the different classifiers, whereas the Confusion matrix is given for the LDA classifier on unseen data, which is deployed for real-time fault detection.

\subsection{Common faults in the SR-30 Fuel supply system}  
In the fuel supply system of SR-30, consists of a series of elements to automate the fuel flow rate in the laboratory  GTE. A fault in the actuator may cause a severe loss in system energy and may cause a total loss of control. Some common faults in servo-actuators, such as lock-in-place (jamming) and actuator offset, are detected in this work.

\begin{figure*}[h!]
	\centering
	\subfigure[]{\includegraphics[height = 6.5 cm, width =7.5 cm]{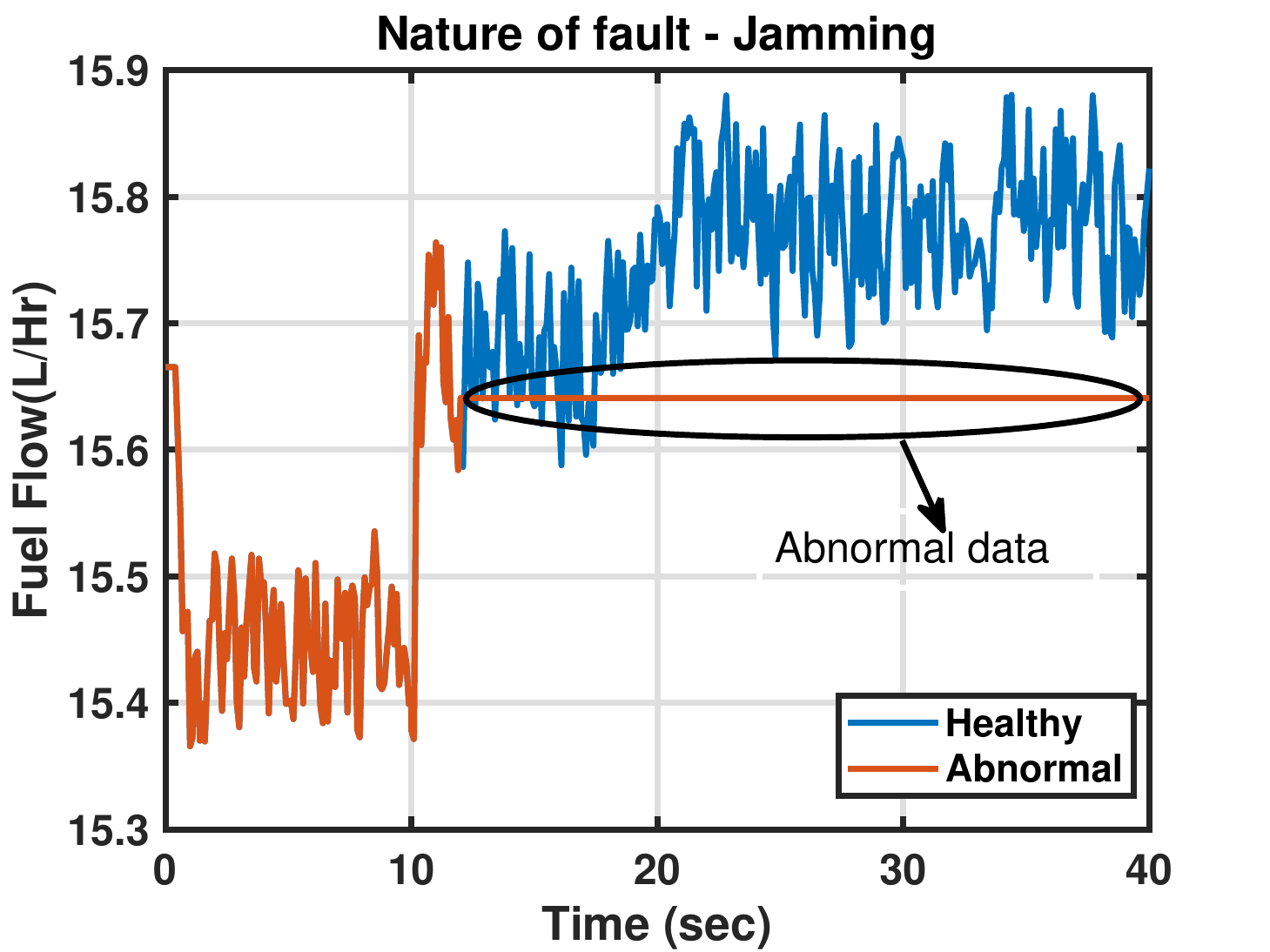}\label{lock1}} 
	\subfigure[]{\includegraphics[height =6.5 cm, width = 7.5 cm]{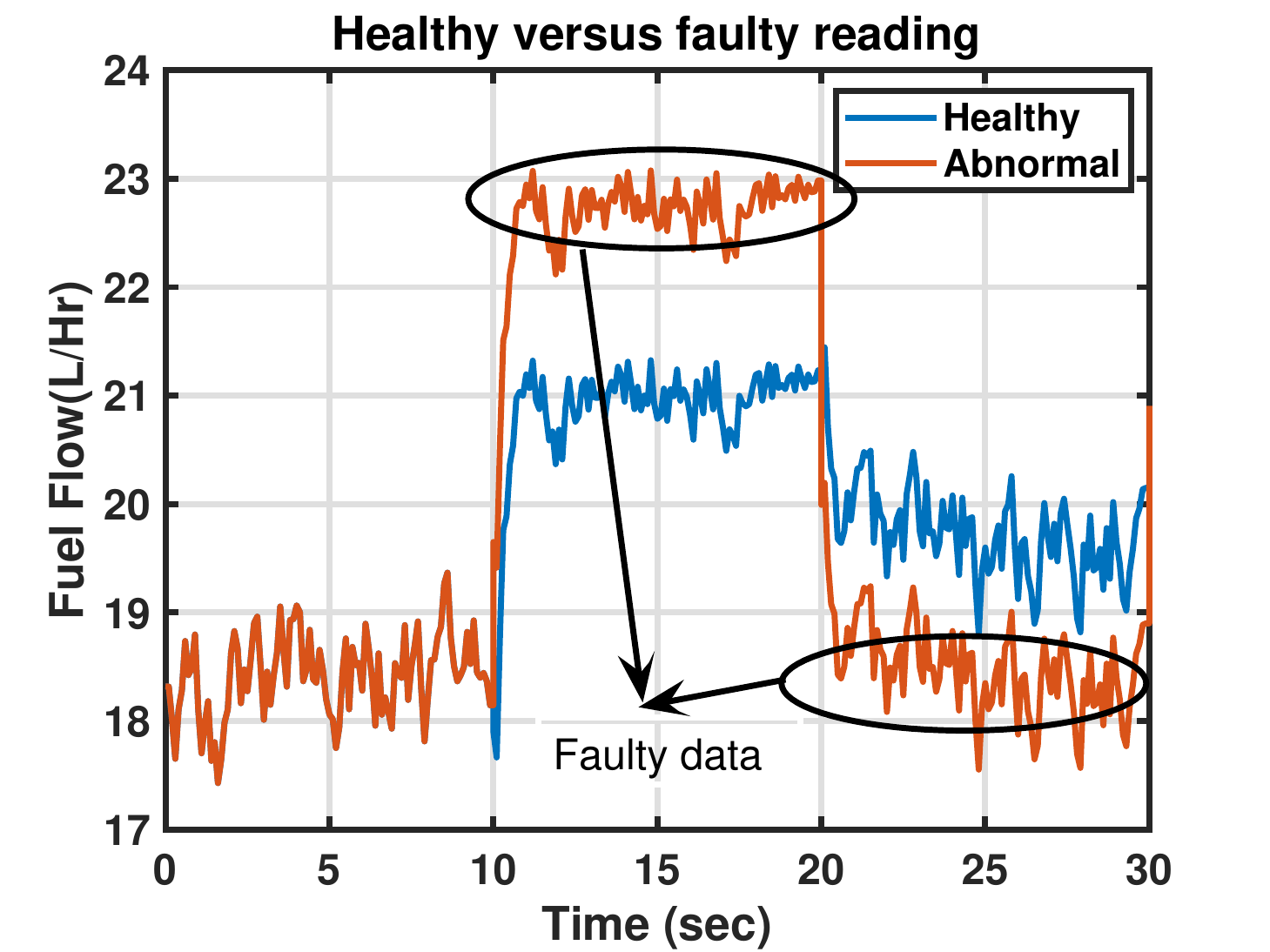}\label{fueldegrade1}} 
	\caption{Nature of faults: \subref{lock1} Lock-in-one place \subref{fueldegrade1} Fuel degradation }
	\label{Fig:actuator_faults}	
\end{figure*}

\noindent
\textbf{\textit{Fault 1: Lock-in-place fault }}
A jamming or lock-in-one place is a system-failure case where the throttling lever is stuck at its current position as in Fig \ref{Fig:actuator_faults}(a). For testing the classifier model, the lock-in-place fault is created in the fuel supply system, and the classification performance is evaluated in terms of the error matrix as shown in Fig.~\ref{Fig:act_fault_CM}.

\noindent
\textbf{\textit{Fault 2: Actuator offset } }  For testing the classifier model, the fault is created by amplifying the output of the fuel supply system from 5\% to 15\%, and the classification performance is evaluated in terms of error matrix as shown in Fig.~\ref{Fig:actuator_faults}(b) and the corresponding confusion matrix (Fig.~\ref{Fig:act_fault_CM}) for sensor degradation.

The confusion matrix shows the degree of misclassification when the trained model is validated on an in-house model of gas turbine engine. Further, the performance metric such as F-1 score and accuracy is compared for all the ML-based classifier for a test dataset and given in Table~\ref{Table:act_offset}. The LDA based classifier delivers better performance in terms of time and accuracy. Although SVM classifier also  shows performance accuracy but the computation time is high.

	\begin{figure}[ht!]
	\centering
	\hspace{1.5 cm}
	\setlength\unitlength{1.5cm}
	\begin{picture}(6,6)
		\multiput(0.1,0.1)(0,1){5}{\line(2,0){4}}
		\multiput(0.1,0.1)(1,0){5}{\line(0,2){4}}
		\put(0.15,0.5){Fault 2}
		\put(1.5,1.5){0}
		\put(2.5,0.5){1}
		
		\put(0.15,2.5){Healthy}
		\put(3.5,0.5){99}
		\put(2.5,1.5){96}
		
		\put(0.15,1.5){Fault 1}
		\put(2.5,2.5){0}
		\put(3.5,1.5){4}
		
		\put(1.13,3.5){Healthy}
		\put(2.13,3.5){Fault 1}
		\put(3.13,3.5){Fault 2}
		
		\put(1.5,0.5){0}
		\put(1.4,2.5){100}
		\put(3.5,2.5){0}
		
		\put (-0.2,1.5){\rotatebox{90}{Actual class}}
		\put(1.5,4.2){Predicted class}
	\end{picture}  
	 \caption{The confusion matrix for actuator faults}
	\label{Fig:act_fault_CM}
\end{figure}

\begin{table}[ht!]
	\begin{center}
		\begin{small}
			\caption{\small Comparison of different classifier for FSS Fault}
			\begin{tabular}{l c c c }   
				\hline
				\textbf{Classifier} $ \downarrow $ &\textbf{ Accuracy } & \textbf{F1 Score} &  \textbf{Training Time}\\
				\hline
				\hline
				\textbf{ LDA}  & 98.6 & 0.98 & 0.56 Sec\\
				\textbf{Linear SVC}  & 99 & 0.989 & 9.58 Sec\\
				\textbf{K-neighbor}  & 93.5 & 0.939 & 2.31 Sec\\
				\textbf{Decision tree} & 97.4 & 0.97 & 0.82 Sec\\
				\hline
			\end{tabular}
			\label{Table:act_offset}
		\end{small}
	\end{center}
\end{table}

\FloatBarrier
\subsection{Single sensor fault diagnosis}
The fault diagnosis technique can be extended to multiple sensor fault detection simultaneously. The faults are created in compressor exit temperature, turbine exit temperature, and compressor exit pressure sensors at different instants by adding/subtracting bias to a healthy operating condition. In the current work, only two sensor failure analysis is presented for the sake of clarity. Two different classifier model is utilized for each sensor assessments.

\begin{figure}[ht!]
	\centering
	\subfigure[]{	\includegraphics[scale = 0.5]{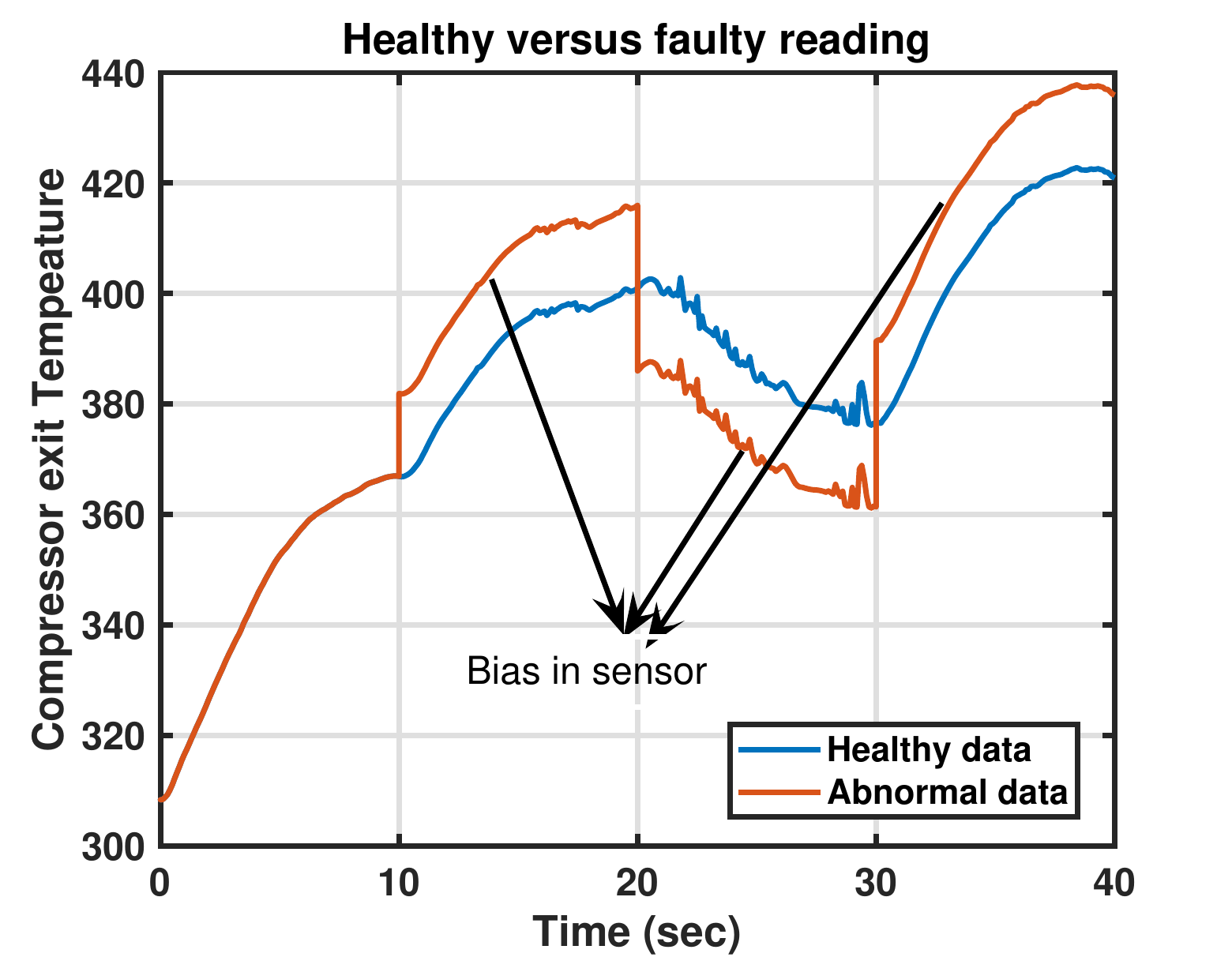} \label{sen_t2}} 
	\subfigure[]{	\setlength\unitlength{1.5cm} 
		\begin{picture}(4,4) 
			\multiput(0.1,0.1)(0,1){4}{\line(2,0){3}}
			\multiput(0.1,0.1)(1,0){4}{\line(0,2){3}}
			\put(0.25,0.5){Faulty}
			\put(1.4,0.5){10.7}
			\put(2.4,0.5){89.3}
			
			\put(0.15,1.5){Healthy}
			\put(1.4,1.5){100}
			\put(2.4,1.5){0}
			
			\put(0.25,2.5){ }
			\put(1.15,2.5){Healthy}
			\put(2.25,2.5){Faulty}
			
			\put (-0.2,1.0){\rotatebox{90}{Actual class}}
			\put(1.0,3.2){Predicted class}
		\end{picture} \label{cmt2}}
	\caption{Nature of faults: \subref{sen_t2} Sensor bias \subref{cmt2}   Confusion matrix faults in $T_2$ sensor}
	\label{Fig:Fault_T2}
\end{figure}

\noindent
\textbf{\textit{Case I: Compressor Exit Temperature $ (T_{2}) $ Sensor Fault Diagnosis:} } For testing the classifier model, the fault is created by amplifying the output of the temperature sensor at the compressor exit by 5\% and the classification performance is evaluated in terms of error matrix. The fault is shown in Fig.~\ref{Fig:Fault_T2}(a). The confusion matrix is obtained for the LDA classifier in Fig.~\ref{Fig:Fault_T2}(b), for detection of the temperature sensor at compressor exit have an accuracy of 94.63\%.

\begin{table}[ht!]
	\begin{center}
		\begin{small}
			\caption{\small Comparison of different classifier for $ (T_{2}) $ sensor}
			\begin{tabular}{l c c c}   
				\hline
				\textbf{Classifier} $ \downarrow $ &\textbf{ Accuracy } & \textbf{F1 Score} &  \textbf{Training Time}\\
				\hline
				\hline
				\textbf{LDA} & 99.8  & 0.985 & 1.47 Sec\\
				\textbf{Linear SVC} & 99.9 & 0.99 & 10.6 Sec\\
				\textbf{K-neighbor} & 98.6 & 0.96 & 3.32 Sec\\
				\textbf{Decision tree} & 81.9 & 0.859 & 0.75 Sec\\
				\hline
			\end{tabular}
			\label{1}
		\end{small}
	\end{center}
\end{table}

\par
\begin{figure}[ht!]
	\centering
	\subfigure[]{\includegraphics[scale = 0.55]{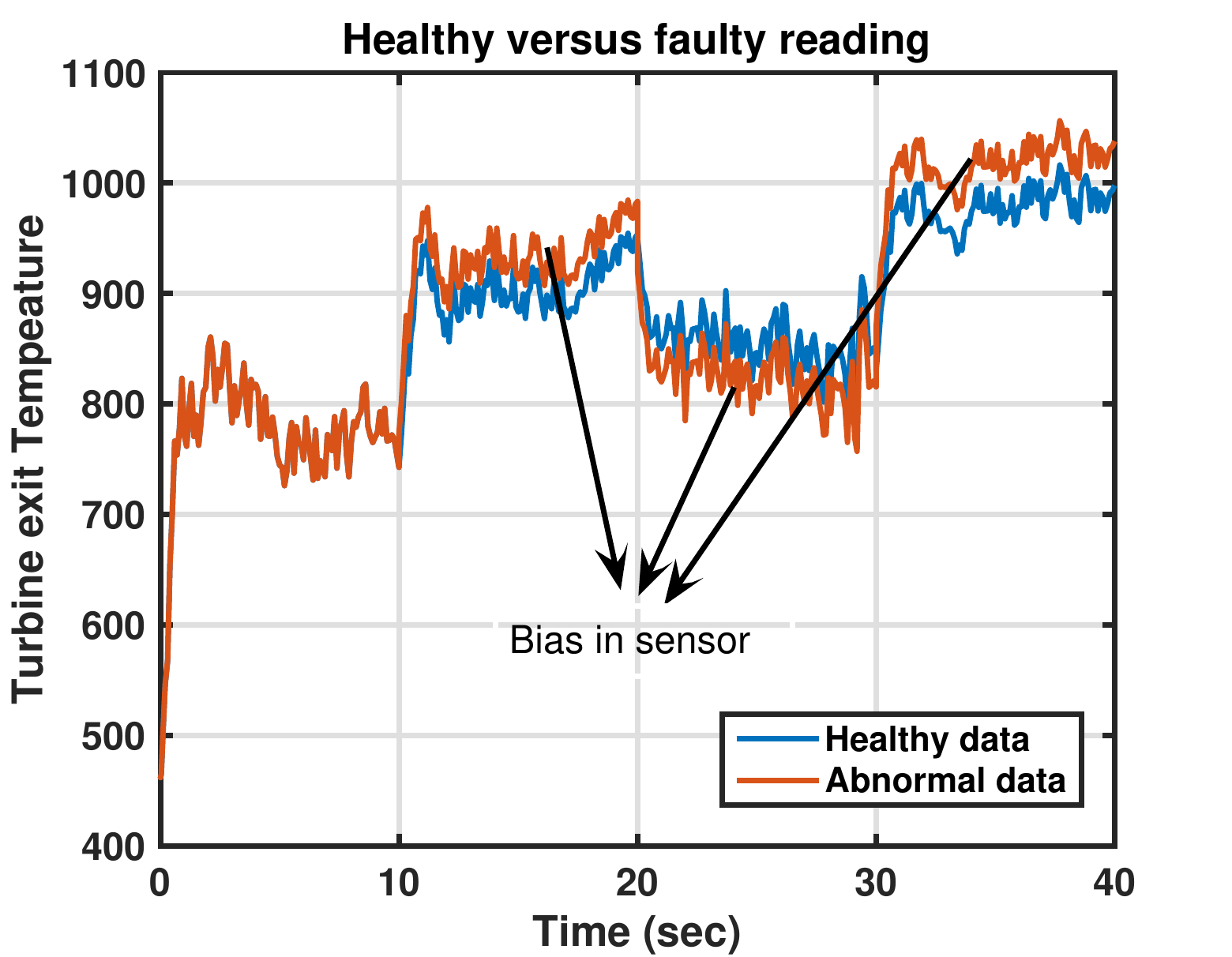} \label{sen_t3}} 
	\subfigure[]{	\setlength\unitlength{1.5cm} 
		\begin{picture}(4,4) 
			\multiput(0.1,0.1)(0,1){4}{\line(2,0){3}}
			\multiput(0.1,0.1)(1,0){4}{\line(0,2){3}}
			\put(0.25,0.5){Faulty}
			\put(1.4,0.5){13.27}
			\put(2.4,0.5){86.73}
			
			\put(0.15,1.5){Healthy}
			\put(1.4,1.5){100}
			\put(2.4,1.5){0}
			
			\put(0.25,2.5){ }
			\put(1.15,2.5){Healthy}
			\put(2.25,2.5){Faulty}
			
			\put (-0.2,1.0){\rotatebox{90}{Actual class}}
			\put(1.0,3.2){Predicted class}
	\end{picture}  \label{cmt3}}
	\caption{Nature of faults: \subref{sen_t3} Sensor bias \subref{cmt3}   Confusion matrix faults in $T_3$ sensor}
	\label{Fig:Fault_T3}
\end{figure}

 From the confusion matrix is obtained for LDA classifier (Fig. \ref{Fig:Fault_T2}),  with an overall accuracy for detection of the temperature sensor at compressor exit fault is found to be 94 \% on on-board testing.\\
\noindent
\textbf{\textit{Case II: Turbine Inlet Temperature $(T_{3}) $ Sensor Fault Diagnosis:} }For testing the classifier model, the fault is created by degrading and amplifying the output of the turbine inlet temperature sensor by 4 to 6 \%, and the classification performance is evaluated in terms of error matrix. The fault in
the sensor is shown in Fig. ~\ref{Fig:Fault_T3}(a). The confusion matrix obtained as in Fig.~\ref{Fig:Fault_T3}(b), is given for LDA Classifier. The accuracy for detection of compressor exit temperature sensor fault is found to be 93\%.

\begin{table}[ht!]
	\begin{center}
		\begin{small}
			\caption{\small Comparison of different classifier for $(T_{3}) $ Sensor}
			\begin{tabular}{l c c c }   
				\hline
				\textbf{Classifier} $ \downarrow $ &\textbf{ Accuracy } & \textbf{F1 Score} &  \textbf{Training Time}\\
				\hline
				\hline
				\textbf{LDA} & 95.5 & 0.964 & 4.02 Sec\\
				\textbf{Linear SVC} & 95.5 & 0.964 & 267.1 Sec\\
				\textbf{K-neighbor} & 92.7 & 0.911 & 5.46 Sec\\
				\textbf{Decision tree} & 74.3 & 0.823 & 1.72 Sec\\
				\hline
			\end{tabular}
			\label{1}
		\end{small}
	\end{center}
\end{table}

\FloatBarrier
\subsection{Multiple sensor fault diagnosis} 
\begin{figure}[ht]
	\centering
	\includegraphics[scale=0.5]{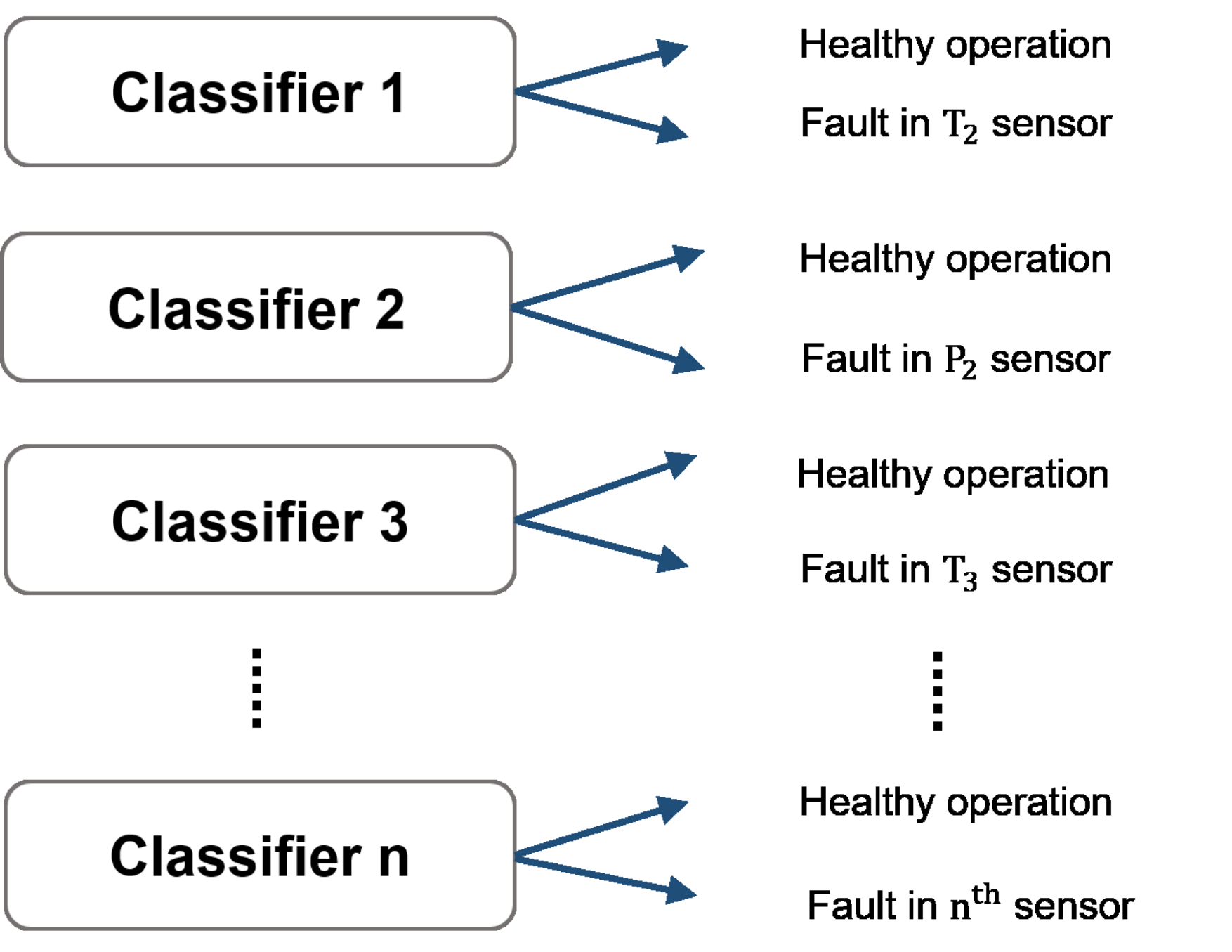} 
	\caption{Multiple model fault diagnosis approach for sensor  failure} 
	\label{Fig:multiple_classifier}
\end{figure}

A multiple model approach is used for multiple sensor fault prediction where each classifier corresponds to a specific sensor of the engine. The on-board fault detection is carried for simultaneous faults in multiple sensors. A dashboard tool ``lamp'' is used as an indicator available in MATLAB/ Simulink toolbox to indicate the faulty measurements. Whenever the FDI model predicts a faulty sample, the lamp shows ``red'' and ``green'' of the lamp indicates a healthy sample. Fig.~\ref{Fig:simulink_multimodel} and \ref{Fig:multiple_classifier} show on-board multiple fault detection interfaces where the in-house simulator is used to analyze the efficacy of the FDI model. Fig 17 gives the confusion matrix for 3 sensors at random for performance evaluation.

\begin{figure}[ht!]
	\centering
	\subfigure[]{	\setlength\unitlength{1.4cm} 
		\begin{picture}(4,4) 
			\multiput(0.1,0.1)(0,1){4}{\line(2,0){3}}
			\multiput(0.1,0.1)(1,0){4}{\line(0,2){3}}
			\put(0.25,0.5){Faulty}
			\put(1.4,0.5){13.9}
			\put(2.4,0.5){86.61}
			
			\put(0.15,1.5){Healthy}
			\put(1.4,1.5){100}
			\put(2.4,1.5){0}
			
			\put(0.25,2.5){ }
			\put(1.15,2.5){Healthy}
			\put(2.25,2.5){Faulty}
			
			\put (-0.2,1.0){\rotatebox{90}{Actual class}}
			\put(1.0,3.2){Predicted class}
		\end{picture}   \label{multi_cmt2}} \hspace{-1.2 cm}
	\subfigure[]{	\setlength\unitlength{1.4cm} 
		\begin{picture}(4,4) 
			\multiput(0.1,0.1)(0,1){4}{\line(2,0){3}}
			\multiput(0.1,0.1)(1,0){4}{\line(0,2){3}}
			\put(0.25,0.5){Faulty}
			\put(1.4,0.5){15.79}
			\put(2.4,0.5){84.21}
			
			\put(0.15,1.5){Healthy}
			\put(1.4,1.5){100}
			\put(2.4,1.5){0}
			
			\put(0.25,2.5){ }
			\put(1.15,2.5){Healthy}
			\put(2.25,2.5){Faulty}
			
			\put (-0.2,1.0){\rotatebox{90}{Actual class}}
			\put(1.0,3.2){Predicted class}
		\end{picture} \label{multi_cmt3}}\hspace{-1.2 cm}
	\subfigure[]{	\setlength\unitlength{1.4cm} 
		\begin{picture}(4,4) 
			\multiput(0.1,0.1)(0,1){4}{\line(2,0){3}}
			\multiput(0.1,0.1)(1,0){4}{\line(0,2){3}}
			\put(0.25,0.5){Faulty}
			\put(1.4,0.5){11.28}
			\put(2.4,0.5){88.24}
			
			\put(0.15,1.5){Healthy}
			\put(1.4,1.5){100}
			\put(2.4,1.5){0}
			
			\put(0.25,2.5){ }
			\put(1.15,2.5){Healthy}
			\put(2.25,2.5){Faulty}
			
			\put (-0.2,1.0){\rotatebox{90}{Actual class}}
			\put(1.0,3.2){Predicted class}
		\end{picture} \label{multi_cmp3}}
	\caption{Nature of faults: \subref{multi_cmt2}  Confusion matrix faults in $T_2$ sensor  \subref{multi_cmt3}   Confusion matrix faults in $T_3$ sensor   \subref{multi_cmp3}   Confusion matrix faults in $T_3$ sensor}
	\label{Fig:multi_Fault_cm}
\end{figure}

\vspace{-3 cm}

\begin{figure}[ht!]
	\centering
	\includegraphics[scale=0.85]{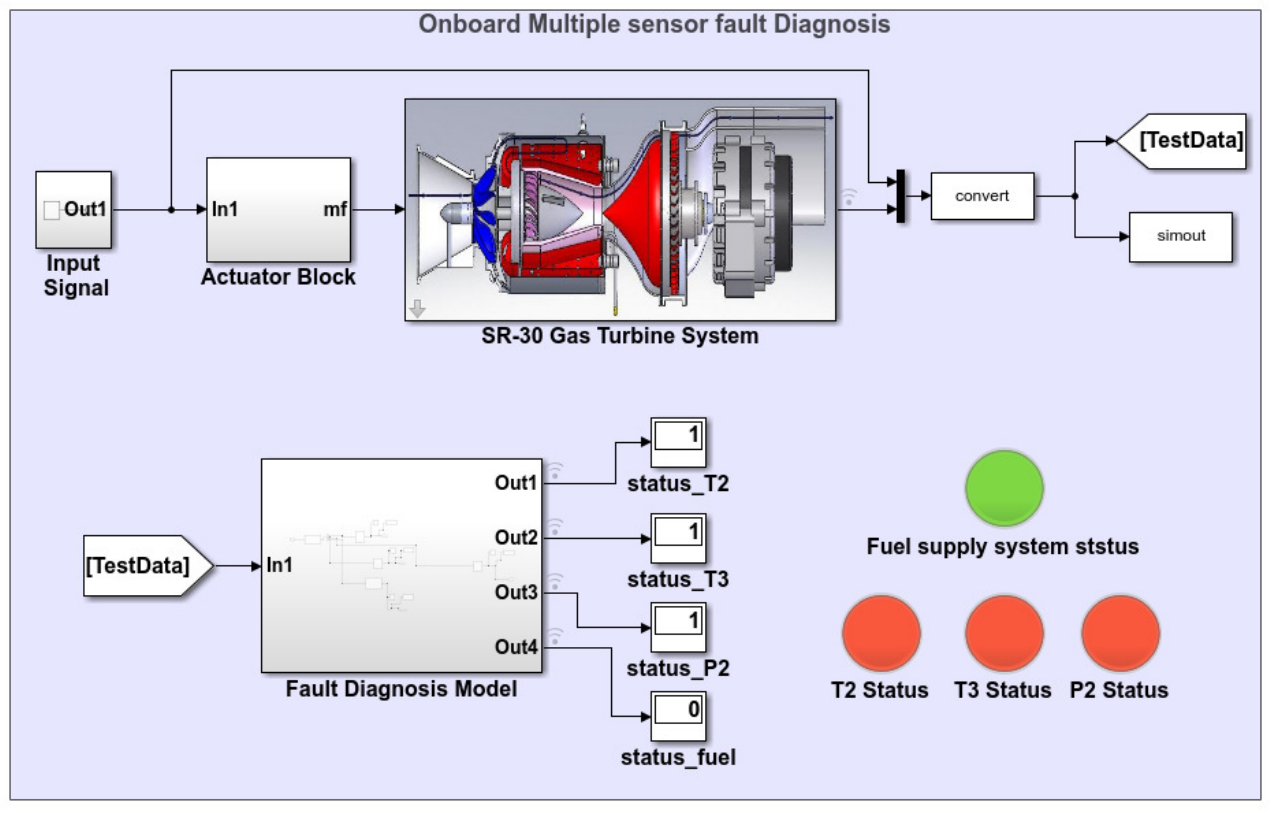} 
	\caption{On-board Multiple sensor fault diagnosis in MATLAB/Simulink}
	\label{Fig:simulink_multimodel}
\end{figure}

\FloatBarrier

From the above simulation results obtained in real-time FDI implementation, the Conclusion can be drawn that with multiple model approach, the fault localization is quite precise, and also the prediction speed improves. The classifiers performance comparison for different fault scenarios shows that the LDA and SVM deliver better performance. However, the training time for the SVM classifier is much greater than that of the LDA classifier, which is critical in real-time diagnosis. Thus for on-board fault detection and localization is carried out by deploying the LDA classifier. The on-board fault detection procedure is carried out successfully. The overall accuracy for multiple faults simultaneously is 94.5\%. The model is able to trap faults successfully for greater than $\pm $3\% deviation from the nominal behavior.

\section{Conclusions}
\label{Conc}
The proposed FDI scheme to diagnose the faults in the fuel supply system and sensor information proved its efficacy in on-board diagnosis and monitoring. The \textit{linear discriminant classifier} is found to give the highest accuracy when compared with other aforementioned classifiers. Although Cubic SVM also delivers a similar performance, the time utilized in training the classifier is much higher ($ \approx $ 13 times) than that for LDA Classifier. The efficacy classification is evaluated by using the confusion matrix. The joint accuracy of the ML-based multiple fault diagnosis model is 93 \% in trapping the faults in the fuel supply system, compressor exit, and turbine inlet temperature sensors. The passive approach makes the model more robust to the failure on which it has been trained.

The component fault and system failure will also be incorporated in future work. Moreover, hybrid techniques may be explored, which involve the advantage of both the model-based and data-driven techniques.

\appendix
\section{Appendix}
\subsection{Confusion Matrix} 
	The performance of any FDI algorithm can be quantified by confusion matrix. A confusion matrix specifies the likelihood of isolating each fault. A typical confusion matrix for for binary classification is shown in Fig \ref{cm}.

	\begin{figure}[ht!]
		\centering
		\hspace{1.5 cm}
		\setlength\unitlength{1.5cm}
		\begin{picture}(4,4)
		\multiput(0.1,0.1)(0,1){4}{\line(2,0){3}}
		\multiput(0.1,0.1)(1,0){4}{\line(0,2){3}}
		\put(0.25,0.5){Faulty}
		\put(1.4,0.5){FN}
		\put(2.4,0.5){TP}
		
		\put(0.15,1.5){Healthy}
		\put(1.4,1.5){TN}
		\put(2.4,1.5){FP}
		
		\put(0.25,2.5){ }
		\put(1.15,2.5){Healthy}
		\put(2.25,2.5){Faulty}
		
		\put (-0.2,1.0){\rotatebox{90}{Actual class}}
		\put(1.0,3.2){Predicted class}
		\end{picture}  
		\caption{Confusion matrix format}
		\label{cm}
	\end{figure}
	
	The interpretation of confusion matrix is as follows:
	\begin{itemize}
		\item \textbf{\textit{True positives (TP):}} We predicted system as faulty, and they are actually faulty.
		\item \textbf{\textit{True negatives (TN):}} We predicted system as healthy, and they are actually healthy.
		\item \textbf{\textit{False positives (FP):}} We predicted system as faulty, but they are not actually faulty. 
		\item \textbf{\textit{False negatives (FN):}} We predicted system as healthy, but they are faulty instead. 
	\end{itemize}
	
	\textbf{Accuracy:} 
	Accuracy is the most intuitive performance measure, and it is simply a ratio of correctly predicted observations to the total observations.  Accuracy is a great measure, but only when you have symmetric datasets. In short, it is the percentage of labels correctly sampled. 
	\[
	Accuracy = \dfrac{TP+TN}{TP+FP+FN+TN}\times 100
	\]
	
	\textbf{$ \mathbf{F1} $ score: } 
	F1 Score score takes both false positives and false negatives into account and approaches 1 as error declines. Intuitively it is not as easy to understand as accuracy, but $ F1 $ is usually more useful than accuracy, especially if you have an uneven class distribution. 
	\[
	F1\,Score = \dfrac{2 TP}{((2 TP) + FP+ FN)}
	\]

\bibliographystyle{plain}	
 \bibliography{reference}

\end{document}